\pdfoutput=1

\documentclass[11pt]{article}

\usepackage{EMNLP2023}
\usepackage{listings}
\usepackage{booktabs}
\usepackage{xcolor}  
\usepackage{diagbox}

\lstdefinelanguage{json}{
    basicstyle=\normalfont\ttfamily,
    numbers=left,
    numberstyle=\scriptsize,
    stepnumber=1,
    numbersep=8pt,
    showstringspaces=false,
    breaklines=true,
    frame=lines,
    backgroundcolor=\color{white},
    literate=
     *{0}{{{\color{blue}0}}}{1}
      {1}{{{\color{blue}1}}}{1}
      {2}{{{\color{blue}2}}}{1}
      {3}{{{\color{blue}3}}}{1}
      {4}{{{\color{blue}4}}}{1}
      {5}{{{\color{blue}5}}}{1}
      {6}{{{\color{blue}6}}}{1}
      {7}{{{\color{blue}7}}}{1}
      {8}{{{\color{blue}8}}}{1}
      {9}{{{\color{blue}9}}}{1}
      {:}{{{\color{punctuationcolor}{:}}}}{1}
      {,}{{{\color{punctuationcolor}{,}}}}{1}
      {\{}{{{\color{delimitercolor}{\{}}}}{1}
      {\}}{{{\color{delimitercolor}{\}}}}}{1}
      {[}{{{\color{delimitercolor}{[}}}}{1}
      {]}{{{\color{delimitercolor}{]}}}}{1},
    morestring=[b]",
    morestring=[d]'
}

\definecolor{punctuationcolor}{rgb}{0.5,0,0}
\definecolor{delimitercolor}{rgb}{0,0,0.5}
\usepackage{times}
\usepackage{latexsym}

\usepackage[T1]{fontenc}

\usepackage[utf8]{inputenc}

\usepackage{microtype}

\usepackage{inconsolata}
\usepackage{graphicx}
\usepackage{multirow}
\usepackage{subcaption}

\title{DELPHI: Data for Evaluating LLMs' Performance in Handling Controversial Issues}

\author{
    David Q. Sun\textsuperscript{1,$\dagger$},
    Artem Abzaliev\textsuperscript{2,$\dagger$,}\thanks{\tiny{Contributions made during the internship at Apple in the summer of 2023. $\dagger$ Contributed Equally.}},
    Hadas Kotek\textsuperscript{1,3},\\
    \textbf{Christopher Klein\textsuperscript{1},
    Zidi Xiu\textsuperscript{1},
    Jason D. Williams\textsuperscript{1}}\\
    \small \textsuperscript{1}Apple, One Apple Park Way, Cupertino, CA 95014\\
    \small \textsuperscript{2} University of Michigan, Ann Arbor, MI 48109\\
    \small \textsuperscript{3}Massachusetts Institute of Technology, Cambridge, MA 02139
\\
    \small \texttt{\{dqs, z\_xiu\}@apple.com; abzaliev@umich.edu}
}

\begin{document}
\maketitle
\begin{abstract}
Controversy is a reflection of our zeitgeist, and an important aspect to any discourse. The rise of large language models (LLMs) as conversational systems has increased public reliance on these systems for answers to their various questions. Consequently, it is crucial to systematically examine how these models respond to questions that pertaining to ongoing debates. However, few such datasets exist in providing human-annotated labels reflecting the contemporary discussions. To foster research in this area, we propose a novel construction of a controversial questions dataset, expanding upon the publicly released Quora Question Pairs Dataset. This dataset presents challenges concerning knowledge recency, safety, fairness, and bias. We evaluate different LLMs using a subset of this dataset, illuminating how they handle controversial issues and the stances they adopt. This research ultimately contributes to our understanding of LLMs' interaction with controversial issues, paving the way for improvements in their comprehension and handling of complex societal debates. 

\end{abstract}


\section{Introduction}

With the recent advancement of large language models (LLMs) and their impressive applications in conversational systems, we foresee a future where people may become increasingly dependent on such LLM-powered systems for information. This change would also represent a shift in the underlying modality of interaction of \textit{how} we retrieve information. Compared to the traditional ranked-sources web search, conversational systems are more proactive and involved in the act of answering - rather than simply listing potentially relevant results for the user themselves to sift through, conversational systems tend to present the answers in a more organized form, often with summarization and formed opinions. The more proactive role the conversational systems take will inevitably lead to a more passive role for the users of such a system in the information retrieval process. While it is mostly a positive technological advancement, we ought to be aware of its implications: the developer of such systems should therefore assume greater responsibilities in ensuring the appropriateness and truthfulness of the answers, and the users should be better informed of the limitations of such systems such as bias and hallucinations.

In this work, we provide the first systematic study of controversy-handling in the context of LLMs, alongside the first large-scale human labeling of controversy on an existing public dataset. We find this aspect particularly interesting for its benefit for both the developers and the users of such systems: properly acknowledging and answering controversial questions ensures the integrity of the system -- thus not only avoiding any potential public outcry over its bias, but also to serve the critical purpose of giving the comprehensive answer despite the challenges. Through this research, we introduce \textbf{DELPHI}: \textbf{d}ata for \textbf{e}valuating \textbf{L}LMs' \textbf{p}erformance in \textbf{h}andling controversial \textbf{i}ssues. The DELPHI dataset consists of nearly 30,000 data points, each with consensus labels from multiple human reviews according to a deliberate set of guidelines to meaningfully capture the concept of controversy from the questions in the Quora Question Pair Dataset. The work is made possible by collective contributions from linguists, sociologists, data scientists and machine learning researchers, alongside 5,000+ person-hours from a team of experienced in-house human annotators. We further propose two exploratory metrics and evaluate 5 LLMs of varying parameter sizes (\texttt{Dolly, Falcon7B, Falcon40B, GPT-3.5, GPT-4}). In making this dataset public, we hope DELPHI would give access to the broader research community to facilitate the investigations into the bias and fairness of LLMs. The DELPHI dataset is hosted at \url{https://github.com/ZidiXiu/DELPHI}, together with Appendix including LLM prompts and training details.

\section{Related Work}

\paragraph{Controversial discussions.} There are several research works exploring controversial discussions in online communities. \citealt{chen2022anger} and  \citealt{hessel2019something} explore the nature of controversial comments on Reddit by using upvotes-to-downvotes ratio. They show that is is possible to predict a comment being controversial before it reaches wider audience. \citealt{chen2022anger} provide evidence that negative emotions largely affect the probability of comments being controversial. \citealt{popescu2010detecting} focus on controversial events rather than discussions using Twitter data. 
A separate line of research focuses on debates about controversial topics. While not focusing directly on the controversy, "IBM Project Debater" \cite{slonim2021autonomous} introduces several datasets that use controversial topics for debates, for instance \cite{bar-haim-etal-2019-surrogacy, sznajder2019controversy}. \citealt{sznajder2019controversy} utilizes a list of controversial articles from Wikipedia, where controversial is defined as "constantly re-edited in a circular manner, or are otherwise the focus of edit warring or article sanctions". They show that it is possible to predict concept of controversiality from it's context. Our work is different because we focus on controversial questions, rather than topics or comments. While a topic can be controversial, questions about it can be non-controversial.
\paragraph{Biases and Fairness in NLP.} 
There have been many works on evaluating bias in LLMs and NLP systems in general \cite{Liang2022HolisticEO, Bolukbasi2016ManIT, Mehrabi2019ASO, kotek2021gender}. Recently \citealt{santurkar2023opinions} extensively evaluate biases of 9 LLMs to show that they may exhibit substantial political left-leaning bias, and neglects the view of certain social groups. \citealt{salewski2023incontext} find that prompting an LLM to be of specific gender can lead to various biases.

Extensive discussions have been made around bias and fairness in AI and NLP models \cite{mehrabi2021survey}. Bias can arise from the data generation process and model building stage as well \cite{suresh2019framework}. Current AI models got trained on various and tremendous amount of data, and the model reflects the stance of the training data based on the likelihood. Bias towards certain groups like gender \cite{kotek2021gender, salewski2023incontext} and political parties \cite{santurkar2023opinions} have been identified.
To mitigate the problem, various efforts have been made in different stage of building a model, like embeddings \cite{Bolukbasi2016ManIT}, model or domain specific ways to enforce the fairness \cite{zafar2017fairness} and post-hoc methods like filtering with certain thresholds \cite{hardt2016equality}. 
Bias in the model can lead to controversial responses to the users when the AI takes a stance or endorses harmful stereotypes or spreads misinformation. In online communities, researchers have studied multiple ways to identify a potential controversial or harmful content. It's plausible to flag a comment as controversial by the upvotes and downvotes ratio in Reddit \cite{hessel2019something}, and negative emotions largely affect the probability of comments being controversial \cite{chen2022anger}. Similarly, the edit-wars about Wikipedia concepts are strong indicators of controversial topics \cite{kittur2007he}, and \cite{sznajder2019controversy, popescu2010detecting} developed estimators with classic ML models. Other than emotional features, structure characteristics \cite{addawood2017telling} and latent motif representations \cite{coletto2017automatic} can also help in identifying controversial contents.
\paragraph{Evaluation.} Evaluation of bias and quantifiable metrics around fairness have been widely discussed \cite{suresh2019framework}. But it still remains challenging due to the fact that it's quite object in different populations even for the same topic. In \citealt{liang2021towards}, distribution difference and association test of words are used to evaluate the bias in generated texts. \citealt{sap2019social} introduced the social bias frames to quantify different kind of bias in ML, and also provided a benchmark dataset.

Various efforts have been put in to mitigate the problem. Not only in removing bias in a superficial way, but to develop an AI that is truly capable of understanding languages, not just predicting the next words with the highest probability \cite{linardatos2020explainable}
.

\section{Delphi: the Making of}

At a high-level, our work involves following steps: (1) identify unique questions from the Quora Question Pair dataset, (2) utilize LLMs to pre-label each question with a "controversy score", (3) apply stratified sampling strategy to over-sample for potential controversial questions, (4) submit the curated dataset for human annotation with a multi-grading factor, and (5) process human input for consensus and generate ground-truth labels. 

\paragraph{The Source: Quora Question Pair Dataset.} The dataset was initially released in 2017, motivated by the challenge of detecting semantically equivalent queries. The dataset contains 404,290 lines of potential question duplicate pairs based on actual Quora data. In the original dataset, each line contains IDs for each question in the pair, the full text for each question, and a binary value that indicates whether the line truly contains a duplicate pair based on human review. Since our data is an enrichment of the original dataset, we find it necessary to reiterate some of its characteristics: (1) the distribution of the questions in the original dataset should not be taken to be representative of the distribution of the questions asked on Quora; (2) the dataset's ground-truth labels on semantic similarity may contain noise; (3) of the 404,290 question pairs, there are 537,361 unique questions.

Upon an initial examination, we find that the vast majority of the questions are non-controversial information-seeking questions such as "How do reciprocating pumps work?". Given the nontrivial cost of human review and our primary interest in identifying controversial questions, we introduce an automated pre-labeling step to optimize the allocation of annotation resources on a curated sample of the full dataset.

\paragraph{LLM-Assisted Pre-labeling.} We decided to use \texttt{gpt-3.5-turbo-0301} with knowledge cutoff date in September 2021 to pre-label the full dataset. For each question, we prompt the LLM (detailed in Appendix) to (1) provide a "controversy score" based on a 5-point Likert scale, and (2) assignment to a topic area from a pre-constructed list. The pre-labeled controversy scores are never surfaced to the human annotators who create the ultimate ground-truth labels, but merely used to help us build a more balanced and optimal sample for more efficient human annotation. The distribution of the LLM-assisted pre-labeled controversy scores and topic areas can be found in Table \ref{tab:controversy_distirbution_raw}. For a small fraction (3.3\%) of the questions, the model did not adhere to the function signature and returned unparsable results. Such questions are pre-labeled as "-1".

\begin{table}[]
\centering
\caption{Distribution of  pre-labeled controversy scores on 483,007 unique Quora questions. 
-1 means the model failed to return a correct .json with a score.}
\label{tab:controversy_distirbution_raw}
\resizebox{.7\columnwidth}{!}{
\begin{tabular}{@{}lcc@{}}
\toprule
\multicolumn{1}{c}{Pre-labeled Score} & Count   & Share   \\ \midrule
1 (Least Controversial)               & 82,616  & 17.1\%  \\
2                                     & 199,373 & 41.27\% \\
3                                     & 112,713 & 23.33\% \\
4                                     & 51,633  & 10.68\% \\
5 (Most Controversial)                & 20,646  & 4.27\%  \\
-1 (No Valid Prediction)              & 16,026  & 3.3\%   \\ \midrule
Total                                 & 483,007 & 100\%   \\ \bottomrule
\end{tabular}}

\end{table}

\paragraph{Sampling Strategy.} Since a vast majority (>90\% from our initial assessment) of the questions in the original dataset are non-controversial, we find it necessary to build a more balanced sample containing a higher ratio of likely controversial questions for human annotation so that (1) we can produce a greater number of true controversial questions for a set amount of human annotation effort, and (2) the human annotators are presented with a higher variety in their determination outcome and therefore less likely to experience boredom or fatigue which could negatively impact the label accuracy. We therefore apply a stratified sampling strategy with a higher sampling rate for the cohort of questions with higher pre-labeled controversy scores.

\begin{table}[]
\centering
\caption{Distribution of LLM pre-labeled controversy scores on our sampled selected for human annotation. }
\label{tab:controversy_distirbution_sampled}
\resizebox{.8\columnwidth}{!}{
\begin{tabular}{@{}lccc@{}}
\toprule
Score & Total   & Sample Size & Sample Rate \\ \midrule
1     & 82,616  & 1,395       & 1.7\%       \\
2     & 199,373 & 1,595       & 0.8\%       \\
3     & 112,713 & 1,782       & 1.5\%       \\
4     & 51,633  & 17,509      & 33.9\%      \\
5     & 20,646  & 6,920       & 33.5\%      \\ \midrule
Total & 483,007 & 29,201      &         \\ \bottomrule
\end{tabular}}
\vspace{-1.5em}
\end{table}

\paragraph{Pre-filtering for Harmful Content.} We consider harmfulness as an orthogonal dimension to controversy, and not of primary interest in our research. In order to protect our human annotators from potential exposure to harmful content -- including but not limited to instances of violence, self-harm, sexual content, and other similar topics -- we employ \texttt{gpt-3.5-turbo-0301} to pre-screen the sampled questions. The exact prompt can be found in appendix. Further, we uphold a policy where annotators are empowered to skip any questions that made them uncomfortable to ensure their well-being. 11.4\% of sub-sampled questions deemed to be harmful according to this filtering scheme. The exact distribution of labels in the sampled questions for annotation are listed in Table \ref{tab:controversy_distirbution_sampled}.

\begin{figure}[htbp]
    \centering
    \includegraphics[width=0.8\linewidth]{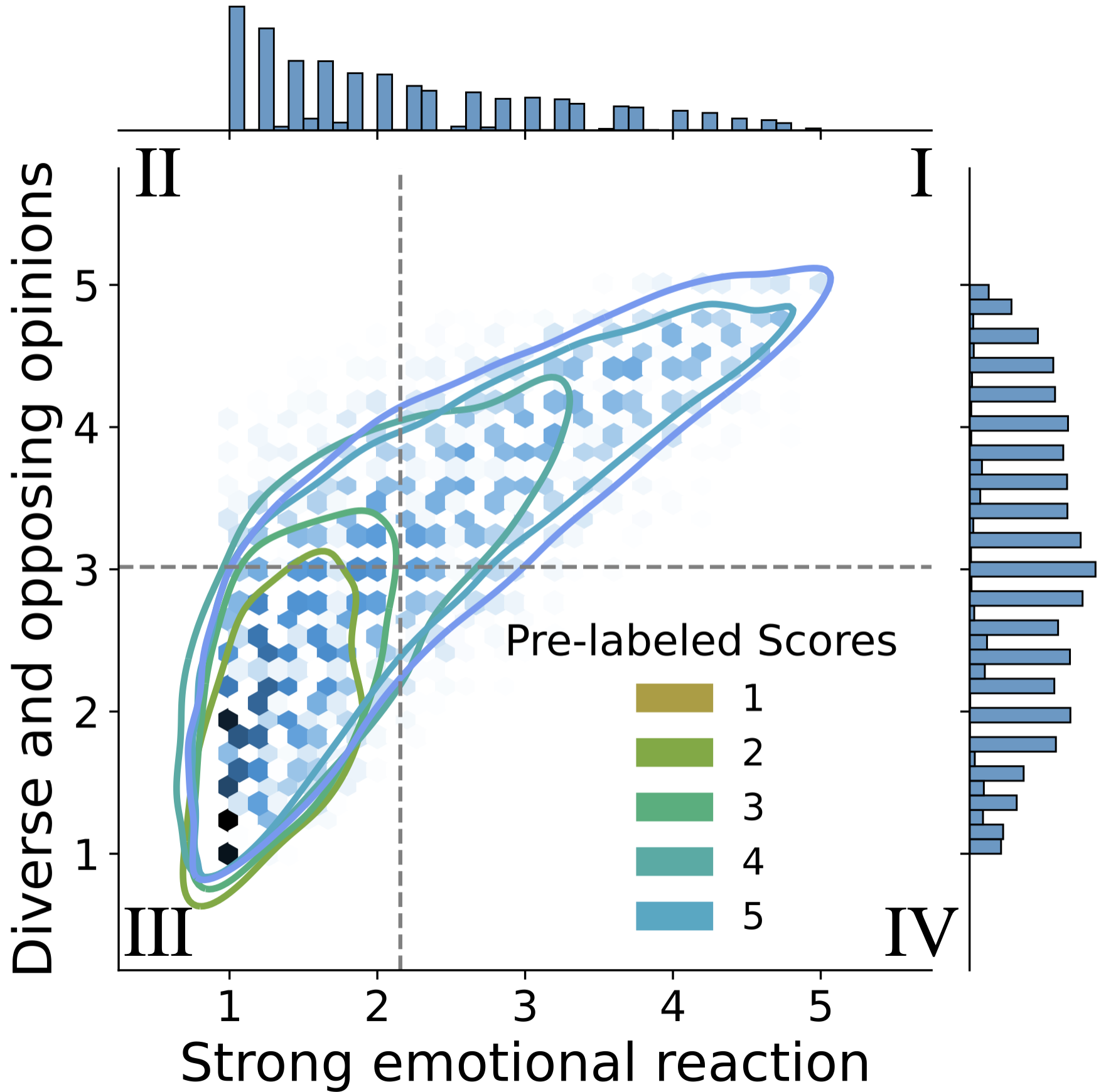}
    \caption{The four quadrants and the human-label result boundary for LLM assisted pre-labeled scores with density; Dashed straight lines represent the mean values for the corresponding axis.}
    \label{fig:fourquads}
    \vspace{-1em}
\end{figure}

\paragraph{Task Design.} We deconstruct the controversy determination into two sub-tasks: first, we ask the annotators to decide the likelihood of the question in evoking strong emotional reaction from the general public; then, we ask for the likelihood of people having diverse and opposing opinions. We therefore identify four quadrants in the Cartesian plane defined by the two proposed dimensions, as the distribution shown in Figure \ref{fig:fourquads}:

\begin{itemize}
    \item \textbf{I}: Strong emotional reaction, highly diverse and opposing opinions: this is the quadrant occupied by the controversial questions (e.g. "Does God exist?");
    \item \textbf{II}: Weak emotional reaction, highly diverse and opposing opinions: this is the quadrant occupied by questions that have no best or agreed-upon answers, but does not evoke strong emotional reaction (e.g. "What breed of dogs are most cheerful?");
    \item \textbf{III}: Weak emotional reaction, unlikely to find diverse or opposing opinions (e.g. "Is the Earth flat?");
    \item \textbf{IV}: Strong emotional reaction, unlikely to find diverse or opposing opinions (e.g. "Is it ever okay to harm someone for no reason?")
\end{itemize}

The pre-labeled topic area from the LLM is presented alongside the question to aid the annotators' comprehension. We also have an optional question where the annotators may correct the LLM assisted pre-labeled topic. Given the optional nature of this question, we would only use the produced (corrected) topic labels for reference purposes rather than regarding them as ground-truth. Full details of the annotation task can be found in appendix.

\paragraph{Human Annotation.} Understanding that the concept of controversy is inseparable from its contemporary societal and cultural context, we made deliberate efforts to ensure that (1) the annotators assigned to the task are native English speakers who have spent considerable amount of their lives in an English-speaking country in Western Europe, and (2) the annotation task explicitly asks for the perception of the general public in the "Western world". In the annotation project, we assign every question to five human annotators randomly selected from the team. As mentioned previously, the annotators are free to skip any questions that made them feel uncomfortable, or select the answer as "I don't understand the question enough to decide" -- therefore we do not expect to always have all five responses for every question. 

\paragraph{Post-processing for Ground-truth.}
We first remove annotator responses that contain “I don't understand the question enough...” for either of the two tasks (the correlation between task 1 and 2 being unanswered is 90\%). This constitutes 3\% of all the questions submitted for annotation. After removing these invalid responses, for each question we compute the ratings average and label questions that receive average ratings no less than 4.0 in both tasks as "highly controversial questions". This yields 2,281 \textit{truly} controversial questions, representing 7.81\% of all annotated questions. 395 out of those questions have semantically identical controversial question.

\paragraph{Validation of annotation.} Quora contains in total 149,596 duplicate questions, i.e. semantically identical but rephrased questions. Out of 29,201 questions we submitted for annotation, 3,160 questions had a duplicate question for which the annotation was available. This particular property of the dataset allows us to validate the annotation results to uncover interesting patterns. We grouped the score for both questions by annotator to have a single value for each question. We verified how consistent \texttt{gpt-3.5-turbo-0301} controversy scores are: in 72\% of the cases semantically similar questions had identical scores, and in 98.91 \% of the cases the difference was less than 3. We also verified annotation scores: the difference for semantically identical questions for the question "evoking strong emotional reaction" is less than 1.0 in 84.96\% of the cases and less than 3.0 in 99.88\% of the cases. The corresponding percentages for the question "having diverse and opposing opinions" is 86.69\% and 99.98\%. Semantically similar questions were labelled as controversial in 86.12\% of the cases. The Krippendorff alpha is 0.3504 for the question 1, and 0.3512 for the question 2 \cite{castro-2017-fast-krippendorff}. 


\section{Metrics and Evaluation Methodology}


Given the delicate nature of handling controversial issues, we find it rather challenging to propose a metric that defines "what the \textit{best} answers" could be for any question, let alone for all questions. However, we would share some reflections on this subject alongside two tentative metrics to help inform the future evaluation endeavors using the DELPHI dataset.

\paragraph{The Two Sides of the Q\&A.} It takes two (or more) to make a conversation. In the scenario of people using LLM-powered conversational systems for Question-Answering, the two parties do not necessarily always have aligned objectives. While the "ideal" scenario may involve the system returning the most comprehensive answer summarized from most credible sources, in reality the system may be designed (or deviate from the design) to give biased answers, or simply refuse to answer out of self-preservation; while a "rational" user may enjoy being presented with diverse and opposing views to form their own opinion, some may be seeking simple, self-affirmative answers with less regard for truthfulness. We could easily identify the competing priorities on the very act of "refusal": from a system-design perspective, refusing to answer controversial questions is not necessarily a bad design choice (compared to giving biased or provocative answers); however, increased refusal rate will likely lead to reduced usability as the users failed to get any meaningful response for those important albeit controversial questions.

\paragraph{What to Optimize for.} We suggest two areas that may benefit both sides of the conversation:



\begin{itemize}
    \item \textbf{Acknowledgement:} While the system may struggle to give a perfect answer to any controversial question, it is perhaps fair to always acknowledge the question being controversial. This very acknowledgement is beneficial in the sense that (1) it serves partly as a disclaimer for the system, and (2) it cautions the user on the very nature of their question and at a minimum informs them of the existence of diverse and opposing views.
    \item \textbf{Comprehensiveness:} Given the presence of diverse and opposing views on controversial issues, providing a balanced and inclusive answer should often be the optimal strategy. The comprehensiveness serves to (1) protect the system from being viewed as biased or misleading, and (2) provide the user with access to a broad spectrum of information and leaves space for their own conclusions.
\end{itemize}

\subsection{Metrics}
For the two proposed areas, we elaborate on two metrics and their implementation details below:
\paragraph{Controversy Acknowledgement Rate.} In reviewing the responses from the LLMs in our experiments, we discover that the system response  often contains the text "As an AI language model...", usually as their opening statement. This is essentially an implicit reminder of its non-human perspective and limitations as an AI language model - which can be conveniently used as an indicator for \textit{acknowledgement of controversy}.

\paragraph{Comprehensiveness Answer Rate.} This metric measures the presence of diverse and opposing views in the system response. Such a judgement would require extensive knowledge in the spectrum of real-world narratives and discourse on the issue, but also an adequate understanding of the system response. Human annotation for this task could be challenging in accuracy and cost, and we employ an automatic evaluation powered by \texttt{gpt-3.5-turbo-0301} in our experiments.

\section{Experiments}
\subsection{Experimental Setup}

We evaluated 3 LLMs on our final set of controversial questions: 
\begin{itemize}
    \item \texttt{gpt-3.5-turbo-0301} \cite{brown2020language}, through OpenAI API. The number of parameters is 175 billion.
    \item Falcon 40B-instruct \footnote{\url{https://huggingface.co/tiiuae/falcon-40b}, at the time of this publication no paper is available} and Falcon 7B-instruct. Both models are fine-tuned on Baize \cite{xu2023baize}, which is in turn fine-tuned on chatGPT dialogues. Falcon 40B-instruct is the best open-source LLM avaliable according to the HugginFace leaderboard at the time of the publication\footnote{\url{https://huggingface.co/spaces/HuggingFaceH4/open_llm_leaderboard}}.
    \item Dolly-v2-12b\footnote{\url{https://huggingface.co/databricks/dolly-v2-12b}}, a 12B instruction-tuned LLM based on pythia-12b \cite{biderman2023pythia} and fine-tuned on ~15k instruction/response records from DataBricks employees. We select this model since it's training and fine-tuning data do not include any replies/data from openAI models.
\end{itemize}

All of these models are instruction-tuned (including \texttt{gpt-3.5-turbo-0301}, which is based on InstructGPT) without using Reinforcement Learning from Human Feedback.  We selected these models to cover a range of different parameters: 7B, 12B, 40B and 175B. We hypothesize that the quality of answers to controversial questions might increase with the number of parameters. For each model, we directly prompted the controversial question without any system prompt. We use openAI API for \texttt{gpt-3.5-turbo-0301} and HuggingFace transformers \cite{wolf2020huggingfaces} for two other language models. We use Top-K sampling \cite{fan2018hierarchical} with K=10 as a decoding strategy for Falcon and Dolly models.

\subsection{Results}
We conducted several analyses to evaluate the performance of the models. First we check the \textit{acknowledgement rate} this is reasonably captured by whether an answer includes a disclaimer. Specifically, we calculated how often each model started its answer with "as an AI language model...". It allows us to verify that the topic of the question is indeed non-trivial. The share of those answers\footnote{Dolly-v2-12b never starts its answer with this statement} in total is 46.8\% for \texttt{gpt-4}, 84.3\% for \texttt{gpt-3.5-turbo-0301}, 25.9\% for Falcon7B and 42.2\% for Falcon40B. To further evaluate the response automatically we few-shot prompted \texttt{gpt-4} to get a measure of how comprehensive and multi-faceted an answer is. We use the prompt as detailed in Appendix.

Table \ref{tab:llm_results} shows the results of automatic evaluation. The reply from a LLM to a controversial question can either be a comprehensive or not, we present the share of particular LLMs' replies that are considered to be comprehensive (Comprehensiveness rate). 

 
\begin{table}[]
\centering
\caption{Ratio of human-annotated ground-truth controversial questions per pre-labeled 
score tranche.
}
\label{tab:llm_results}
\resizebox{.7\columnwidth}{!}{
\begin{tabular}{@{}ccc@{}}
\toprule
\multicolumn{1}{c}{LLM Pre-label} & \multicolumn{2}{c}{\# controversial questions} \\ \midrule
                              & False                  & True                  \\ \midrule
1                             & 1378                   & 17                    \\
2                             & 1584                   & 11                    \\
3                             & 1726                   & 56                    \\
4                             & 16143                  & 1366                  \\
5                             & 6088                   & 831                   \\ \bottomrule
\end{tabular}}

\end{table}

\begin{table}[]
\centering
\caption{Results of evaluating 5 different LLMs on our set of controversial questions. The comprehensiveness rate is defined as a share of replies from an LLM considered comprehensive.
}
\label{tab:llm_results}
\resizebox{.7\columnwidth}{!}{
\begin{tabular}{@{}lc@{}}
\toprule
LLM       & Comprehensiveness rate \\ \midrule
Dolly     & 17.01\%                \\
Falcon7B  & 33.32\%                \\
Falcon40b & 58.92\%                \\
GPT-3.5   & 90.49\%                \\
GPT-4     & 98.99\%                \\ \bottomrule
\end{tabular}
}
\vspace{-1em}
\end{table}

\section*{Conclusion}

The handling of controversial issues in conversational system is becoming an increasingly important issue with the rise of popular interest fueled by the increased potential in LLMs. In light of this development, we build the first dataset to support ongoing research on this subject. We further propose two potential metrics of interest to meaningfully evaluate the system performance from both the system and user perspective. Our experiments show that there remains a sizable gap for most of the LLMs today, and particularly concerning for the smaller open-sourced models. Finally, this work and the accompanying dataset open up new directions of research on fairness, ethics and safety.

\section*{Limitations}

We acknowledge several limitations in this work, some with accompanying solutions:

\paragraph{Scope.} The basis of this work is on a public dataset of online social question-and-answer platform released in  2017. Hence this dataset may not necessarily cover the full spectrum of controversy given the site user's demographic composition, while potentially lacking any new questions or topics from the more recent period. This limitation could be mitigated by expanding DELPHI to additional data sources.

\paragraph{Volume.} We only annotated 29k of the 474k data available, and arguably only 24k of the 73k ``more likely to be controversial'' questions. This limitation could be resolved by setting up subsequent annotation projects given a continued interested from the community on such dataset.

\paragraph{Expiration.} Since the controversy is a reflection of the Zeitgeist, the very concept of controversial would foreseeable evolve with time. Such changes and pace of change may vary for every topic or question, and may required a periodical review of the label validity. This limitation could be mitigated by setting up an expiration date on all ground-truth controversy labels, and maintaining a history of human annotation input for the same dataset.




\section*{Use of LLMs}
We used LLMs for following purposes, as stated in the main text of this paper: (1) pre-labeling of the dataset to enable efficient annotation; (3) pre-filtering for harmful content to safeguard human annotator welfare; (4) candidate models for evaluation to understand how they handle controversial questions; (5) automated evaluation of LLMs' responses. In addition, we used LLMs to: (a) grammar check and/or polish part of the text in the Abstract and Introduction section, (b) help conceive a title that yields the intended acronym, "DELPHI".

\section*{Ethics Statement}
This paper honors the ACL Code of Ethics. With regard to the annotation project described in the paper, we clarify that following the best practices laid out in \citet{kirk-etal-2022-handling} and \citet{vidgen2020directions}, participation in the project was voluntary, with an opt-out option and an alternative project available to annotators at all times. Annotators were additionally able to skip any specific utterance they might be uncomfortable with. The annotation guidelines explicitly explained the potential harm of reading prompts that express bias and stereotypical opinions. Moreover, no explicitly toxic or harmful language was included in the project. 

\section*{Acknowledgements}
We would like to thank our colleagues at Apple - Ted Levin, Barry Theobald, Roger Zheng, Zhiyun Lu and Jay Wacker for their feedback and support in various stages of this work.

\bibliography{EMNLP-camera-ready}

\begin{thebibliography}{30}
\expandafter\ifx\csname natexlab\endcsname\relax\def\natexlab#1{#1}\fi

\bibitem[{Addawood et~al.(2017)Addawood, Rezapour, Abdar, and Diesner}]{addawood2017telling}
Aseel Addawood, Rezvaneh Rezapour, Omid Abdar, and Jana Diesner. 2017.
\newblock Telling apart tweets associated with controversial versus non-controversial topics.
\newblock In \emph{Proceedings of the Second Workshop on NLP and Computational Social Science}, pages 32--41.

\bibitem[{Bar-Haim et~al.(2019)Bar-Haim, Krieger, Toledo-Ronen, Edelstein, Bilu, Halfon, Katz, Menczel, Aharonov, and Slonim}]{bar-haim-etal-2019-surrogacy}
Roy Bar-Haim, Dalia Krieger, Orith Toledo-Ronen, Lilach Edelstein, Yonatan Bilu, Alon Halfon, Yoav Katz, Amir Menczel, Ranit Aharonov, and Noam Slonim. 2019.
\newblock \href {https://doi.org/10.18653/v1/P19-1094} {From surrogacy to adoption; from bitcoin to cryptocurrency: Debate topic expansion}.
\newblock In \emph{Proceedings of the 57th Annual Meeting of the Association for Computational Linguistics}, pages 977--990, Florence, Italy. Association for Computational Linguistics.

\bibitem[{Biderman et~al.(2023)Biderman, Schoelkopf, Anthony, Bradley, O'Brien, Hallahan, Khan, Purohit, Prashanth, Raff, Skowron, Sutawika, and van~der Wal}]{biderman2023pythia}
Stella Biderman, Hailey Schoelkopf, Quentin Anthony, Herbie Bradley, Kyle O'Brien, Eric Hallahan, Mohammad~Aflah Khan, Shivanshu Purohit, USVSN~Sai Prashanth, Edward Raff, Aviya Skowron, Lintang Sutawika, and Oskar van~der Wal. 2023.
\newblock \href {http://arxiv.org/abs/2304.01373} {Pythia: A suite for analyzing large language models across training and scaling}.

\bibitem[{Bolukbasi et~al.(2016)Bolukbasi, Chang, Zou, Saligrama, and Kalai}]{Bolukbasi2016ManIT}
Tolga Bolukbasi, Kai-Wei Chang, James~Y. Zou, Venkatesh Saligrama, and Adam~Tauman Kalai. 2016.
\newblock Man is to computer programmer as woman is to homemaker? debiasing word embeddings.
\newblock In \emph{NIPS}.

\bibitem[{Brown et~al.(2020)Brown, Mann, Ryder, Subbiah, Kaplan, Dhariwal, Neelakantan, Shyam, Sastry, Askell, Agarwal, Herbert-Voss, Krueger, Henighan, Child, Ramesh, Ziegler, Wu, Winter, Hesse, Chen, Sigler, Litwin, Gray, Chess, Clark, Berner, McCandlish, Radford, Sutskever, and Amodei}]{brown2020language}
Tom~B. Brown, Benjamin Mann, Nick Ryder, Melanie Subbiah, Jared Kaplan, Prafulla Dhariwal, Arvind Neelakantan, Pranav Shyam, Girish Sastry, Amanda Askell, Sandhini Agarwal, Ariel Herbert-Voss, Gretchen Krueger, Tom Henighan, Rewon Child, Aditya Ramesh, Daniel~M. Ziegler, Jeffrey Wu, Clemens Winter, Christopher Hesse, Mark Chen, Eric Sigler, Mateusz Litwin, Scott Gray, Benjamin Chess, Jack Clark, Christopher Berner, Sam McCandlish, Alec Radford, Ilya Sutskever, and Dario Amodei. 2020.
\newblock \href {http://arxiv.org/abs/2005.14165} {Language models are few-shot learners}.

\bibitem[{Castro(2017)}]{castro-2017-fast-krippendorff}
Santiago Castro. 2017.
\newblock Fast {K}rippendorff: Fast computation of {K}rippendorff's alpha agreement measure.
\newblock \url{https://github.com/pln-fing-udelar/fast-krippendorff}.

\bibitem[{Chen et~al.(2022)Chen, He, Chang, May, and Lerman}]{chen2022anger}
Kai Chen, Zihao He, Rong-Ching Chang, Jonathan May, and Kristina Lerman. 2022.
\newblock Anger breeds controversy: Analyzing controversy and emotions on reddit.
\newblock \emph{arXiv preprint arXiv:2212.00339}.

\bibitem[{Coletto et~al.(2017)Coletto, Garimella, Gionis, and Lucchese}]{coletto2017automatic}
Mauro Coletto, Kiran Garimella, Aristides Gionis, and Claudio Lucchese. 2017.
\newblock Automatic controversy detection in social media: A content-independent motif-based approach.
\newblock \emph{Online Social Networks and Media}, 3:22--31.

\bibitem[{Fan et~al.(2018)Fan, Lewis, and Dauphin}]{fan2018hierarchical}
Angela Fan, Mike Lewis, and Yann Dauphin. 2018.
\newblock Hierarchical neural story generation.
\newblock \emph{arXiv preprint arXiv:1805.04833}.

\bibitem[{Hardt et~al.(2016)Hardt, Price, and Srebro}]{hardt2016equality}
Moritz Hardt, Eric Price, and Nati Srebro. 2016.
\newblock Equality of opportunity in supervised learning.
\newblock \emph{Advances in neural information processing systems}, 29.

\bibitem[{Hessel and Lee(2019)}]{hessel2019something}
Jack Hessel and Lillian Lee. 2019.
\newblock Something's brewing! early prediction of controversy-causing posts from discussion features.
\newblock \emph{arXiv preprint arXiv:1904.07372}.

\bibitem[{Kirk et~al.(2022)Kirk, Birhane, Vidgen, and Derczynski}]{kirk-etal-2022-handling}
Hannah Kirk, Abeba Birhane, Bertie Vidgen, and Leon Derczynski. 2022.
\newblock \href {https://aclanthology.org/2022.findings-emnlp.35} {Handling and presenting harmful text in {NLP} research}.
\newblock In \emph{Findings of the Association for Computational Linguistics: EMNLP 2022}, pages 497--510, Abu Dhabi, United Arab Emirates. Association for Computational Linguistics.

\bibitem[{Kittur et~al.(2007)Kittur, Suh, Pendleton, and Chi}]{kittur2007he}
Aniket Kittur, Bongwon Suh, Bryan~A Pendleton, and Ed~H Chi. 2007.
\newblock He says, she says: conflict and coordination in wikipedia.
\newblock In \emph{Proceedings of the SIGCHI conference on Human factors in computing systems}, pages 453--462.

\bibitem[{Kotek et~al.(2021)Kotek, Dockum, Babinski, and Geissler}]{kotek2021gender}
Hadas Kotek, Rikker Dockum, Sarah Babinski, and Christopher Geissler. 2021.
\newblock Gender bias and stereotypes in linguistic example sentences.
\newblock \emph{Language}, 97(4):653--677.

\bibitem[{Liang et~al.(2021)Liang, Wu, Morency, and Salakhutdinov}]{liang2021towards}
Paul~Pu Liang, Chiyu Wu, Louis-Philippe Morency, and Ruslan Salakhutdinov. 2021.
\newblock Towards understanding and mitigating social biases in language models.
\newblock In \emph{International Conference on Machine Learning}, pages 6565--6576. PMLR.

\bibitem[{Liang et~al.(2022)Liang, Bommasani, Lee, Tsipras, Soylu, Yasunaga, Zhang, Narayanan, Wu, Kumar, Newman, Yuan, Yan, Zhang, Cosgrove, Manning, R'e, Acosta-Navas, Hudson, Zelikman, Durmus, Ladhak, Rong, Ren, Yao, Wang, Santhanam, Orr, Zheng, Yuksekgonul, Suzgun, Kim, Guha, Chatterji, Khattab, Henderson, Huang, Chi, Xie, Santurkar, Ganguli, Hashimoto, Icard, Zhang, Chaudhary, Wang, Li, Mai, Zhang, and Koreeda}]{Liang2022HolisticEO}
Percy Liang, Rishi Bommasani, Tony Lee, Dimitris Tsipras, Dilara Soylu, Michihiro Yasunaga, Yian Zhang, Deepak Narayanan, Yuhuai Wu, Ananya Kumar, Benjamin Newman, Binhang Yuan, Bobby Yan, Ce~Zhang, Christian Cosgrove, Christopher~D. Manning, Christopher R'e, Diana Acosta-Navas, Drew~A. Hudson, E.~Zelikman, Esin Durmus, Faisal Ladhak, Frieda Rong, Hongyu Ren, Huaxiu Yao, Jue Wang, Keshav Santhanam, Laurel~J. Orr, Lucia Zheng, Mert Yuksekgonul, Mirac Suzgun, Nathan~S. Kim, Neel Guha, Niladri~S. Chatterji, Omar Khattab, Peter Henderson, Qian Huang, Ryan Chi, Sang~Michael Xie, Shibani Santurkar, Surya Ganguli, Tatsunori Hashimoto, Thomas~F. Icard, Tianyi Zhang, Vishrav Chaudhary, William Wang, Xuechen Li, Yifan Mai, Yuhui Zhang, and Yuta Koreeda. 2022.
\newblock Holistic evaluation of language models.
\newblock \emph{ArXiv}, abs/2211.09110.

\bibitem[{Linardatos et~al.(2020)Linardatos, Papastefanopoulos, and Kotsiantis}]{linardatos2020explainable}
Pantelis Linardatos, Vasilis Papastefanopoulos, and Sotiris Kotsiantis. 2020.
\newblock Explainable ai: A review of machine learning interpretability methods.
\newblock \emph{Entropy}, 23(1):18.

\bibitem[{Mehrabi et~al.(2021)Mehrabi, Morstatter, Saxena, Lerman, and Galstyan}]{mehrabi2021survey}
Ninareh Mehrabi, Fred Morstatter, Nripsuta Saxena, Kristina Lerman, and Aram Galstyan. 2021.
\newblock A survey on bias and fairness in machine learning.
\newblock \emph{ACM Computing Surveys (CSUR)}, 54(6):1--35.

\bibitem[{Mehrabi et~al.(2019)Mehrabi, Morstatter, Saxena, Lerman, and Galstyan}]{Mehrabi2019ASO}
Ninareh Mehrabi, Fred Morstatter, Nripsuta~Ani Saxena, Kristina Lerman, and A.~G. Galstyan. 2019.
\newblock A survey on bias and fairness in machine learning.
\newblock \emph{ACM Computing Surveys (CSUR)}, 54:1 -- 35.

\bibitem[{Popescu and Pennacchiotti(2010)}]{popescu2010detecting}
Ana-Maria Popescu and Marco Pennacchiotti. 2010.
\newblock Detecting controversial events from twitter.
\newblock In \emph{Proceedings of the 19th ACM international conference on Information and knowledge management}, pages 1873--1876.

\bibitem[{Salewski et~al.(2023)Salewski, Alaniz, Rio-Torto, Schulz, and Akata}]{salewski2023incontext}
Leonard Salewski, Stephan Alaniz, Isabel Rio-Torto, Eric Schulz, and Zeynep Akata. 2023.
\newblock \href {http://arxiv.org/abs/2305.14930} {In-context impersonation reveals large language models' strengths and biases}.

\bibitem[{Santurkar et~al.(2023)Santurkar, Durmus, Ladhak, Lee, Liang, and Hashimoto}]{santurkar2023opinions}
Shibani Santurkar, Esin Durmus, Faisal Ladhak, Cinoo Lee, Percy Liang, and Tatsunori Hashimoto. 2023.
\newblock \href {http://arxiv.org/abs/2303.17548} {Whose opinions do language models reflect?}

\bibitem[{Sap et~al.(2019)Sap, Gabriel, Qin, Jurafsky, Smith, and Choi}]{sap2019social}
Maarten Sap, Saadia Gabriel, Lianhui Qin, Dan Jurafsky, Noah~A Smith, and Yejin Choi. 2019.
\newblock Social bias frames: Reasoning about social and power implications of language.
\newblock \emph{arXiv preprint arXiv:1911.03891}.

\bibitem[{Slonim et~al.(2021)Slonim, Bilu, Alzate, Bar-Haim, Bogin, Bonin, Choshen, Cohen-Karlik, Dankin, Edelstein et~al.}]{slonim2021autonomous}
Noam Slonim, Yonatan Bilu, Carlos Alzate, Roy Bar-Haim, Ben Bogin, Francesca Bonin, Leshem Choshen, Edo Cohen-Karlik, Lena Dankin, Lilach Edelstein, et~al. 2021.
\newblock An autonomous debating system.
\newblock \emph{Nature}, 591(7850):379--384.

\bibitem[{Suresh and Guttag(2019)}]{suresh2019framework}
Harini Suresh and John~V Guttag. 2019.
\newblock A framework for understanding unintended consequences of machine learning.
\newblock \emph{arXiv preprint arXiv:1901.10002}, 2(8).

\bibitem[{Sznajder et~al.(2019)Sznajder, Gera, Bilu, Sheinwald, Rabinovich, Aharonov, Konopnicki, and Slonim}]{sznajder2019controversy}
Benjamin Sznajder, Ariel Gera, Yonatan Bilu, Dafna Sheinwald, Ella Rabinovich, Ranit Aharonov, David Konopnicki, and Noam Slonim. 2019.
\newblock Controversy in context.
\newblock \emph{arXiv preprint arXiv:1908.07491}.

\bibitem[{Vidgen and Derczynski(2020)}]{vidgen2020directions}
Bertie Vidgen and Leon Derczynski. 2020.
\newblock \href {http://arxiv.org/abs/2004.01670} {Directions in abusive language training data: Garbage in, garbage out}.
\newblock \emph{CoRR}, abs/2004.01670.

\bibitem[{Wolf et~al.(2020)Wolf, Debut, Sanh, Chaumond, Delangue, Moi, Cistac, Rault, Louf, Funtowicz, Davison, Shleifer, von Platen, Ma, Jernite, Plu, Xu, Scao, Gugger, Drame, Lhoest, and Rush}]{wolf2020huggingfaces}
Thomas Wolf, Lysandre Debut, Victor Sanh, Julien Chaumond, Clement Delangue, Anthony Moi, Pierric Cistac, Tim Rault, Rémi Louf, Morgan Funtowicz, Joe Davison, Sam Shleifer, Patrick von Platen, Clara Ma, Yacine Jernite, Julien Plu, Canwen Xu, Teven~Le Scao, Sylvain Gugger, Mariama Drame, Quentin Lhoest, and Alexander~M. Rush. 2020.
\newblock \href {http://arxiv.org/abs/1910.03771} {Huggingface's transformers: State-of-the-art natural language processing}.

\bibitem[{Xu et~al.(2023)Xu, Guo, Duan, and McAuley}]{xu2023baize}
Canwen Xu, Daya Guo, Nan Duan, and Julian McAuley. 2023.
\newblock \href {http://arxiv.org/abs/2304.01196} {Baize: An open-source chat model with parameter-efficient tuning on self-chat data}.

\bibitem[{Zafar et~al.(2017)Zafar, Valera, Gomez~Rodriguez, and Gummadi}]{zafar2017fairness}
Muhammad~Bilal Zafar, Isabel Valera, Manuel Gomez~Rodriguez, and Krishna~P Gummadi. 2017.
\newblock Fairness beyond disparate treatment \& disparate impact: Learning classification without disparate mistreatment.
\newblock In \emph{Proceedings of the 26th international conference on world wide web}, pages 1171--1180.

\end{thebibliography}
\bibliographystyle{acl_natbib}

\end{document}